\documentclass[letterpaper, 10 pt, conference]{ieeeconf}  %
\usepackage{caption}

\IEEEoverridecommandlockouts                            
\let\labelindent\relax

\usepackage{graphicx}
\usepackage{booktabs}
\usepackage{float}
\usepackage{cite}
\usepackage{amssymb}
\usepackage{xcolor}
\usepackage{svg}
\usepackage{amsmath}
\usepackage{fancyhdr}
\usepackage{multirow} 
\usepackage{xspace}
\usepackage{tabularx}
\usepackage{enumitem}
\usepackage{subfigure}
\usepackage{hyperref} 
\usepackage[switch]{lineno}

\newcommand{\thickhline}{%
    \noalign {\ifnum 0=`}\fi \hrule height 1pt
    \futurelet \reserved@a \@xhline
}
\newcolumntype{C}{>{\centering\arraybackslash}1.2}

\DeclareRobustCommand\onedot{\futurelet\@let@token\@onedot}
\def\onedot{\ifx\@let@token.\else.\null\fi\xspace}
\def\eg{\emph{e.g}\onedot}

\def\ie{\emph{i.e}\onedot}

\def\etc{\emph{etc}\onedot} 

\def\wrt{w.r.t\onedot}

\makeatletter
\let\NAT@parse\undefined
\makeatother
\usepackage{hyperref} 

\definecolor{Gray}{gray}{0.9}
\definecolor{somegray}{rgb}{0.5, 0.5, 0.5}
\newcommand{\darkgrayed}[1]{\textcolor{somegray}{#1}}

\makeatletter
\newcommand*\titleheader[1]{\gdef\@titleheader{#1}}
\AtBeginDocument{%
  \let\st@red@title\@title
  \def\@title{%
    \vskip-3em
    \bgroup\normalfont\large\centering\@titleheader\par\egroup
    \vskip1.5em\st@red@title}
}
\makeatother

\titleheader{ \darkgrayed{This paper has been accepted for publication at the \\ IEEE International Conference on Robotics and Automation, Yokohama, 2024 
\copyright IEEE} }

\title{\LARGE \bf ESP: Extro-Spective Prediction for Long-term Behavior Reasoning in Emergency Scenarios}

\author{Dingrui Wang$^{1,4}$, Zheyuan Lai$^{1}$, Yuda Li$^{1}$, Yi Wu$^{2}$, Yuexin Ma$^{3}$, Johannes Betz$^{4}$, \\ Ruigang Yang$^{1}$, \emph{Fellow,  IEEE}, Wei Li$^{1}$
\thanks{*The study was supported by the Natural Science Foundation of Jiangsu Province (BK20210600)}%
\thanks{$^{1}$Inceptio Technology, Shanghai 200082, China, 
        {\tt\footnotesize dingrui.wang, zheyuan.lai, yuda.li, yang.ruigang, wei.li @inceptio.ai}}%
\thanks{$^{2}$Nanjing University of Posts and Telecommunications, Nanjing 210023, China, 
        {\tt\footnotesize yiw@njupt.edu.cn}}%
\thanks{$^{3}$ShanghaiTech University, Shanghai 200120, China, 
        {\tt\footnotesize mayuexin@shanghaitech.edu.cn}}%
\thanks{$^{4}$The authors are with the Professorship of Autonomous Vehicle Systems, Technical University of Munich, 85748 Garching, Germany; Munich Institute of Robotics and Machine Intelligence (MIRMI), 
        {\tt\footnotesize dingrui.wang@hotmail.com, johannes.betz@tum.de}}%
}
\begin{document}
\maketitle

\thispagestyle{empty}
\pagestyle{empty}

\begin{abstract} 
Emergent-scene safety is the key milestone for fully autonomous driving, and reliable on-time prediction is essential to maintain safety in emergency scenarios. However, these emergency scenarios are long-tailed and hard to collect, which restricts the system from getting reliable predictions. In this paper, we build a new dataset, which aims at the long-term prediction with the inconspicuous state variation in history for the emergency event, named the Extro-Spective Prediction (ESP) problem. Based on the proposed dataset, a flexible feature encoder for ESP is introduced to various prediction methods as a seamless plug-in, and its consistent performance improvement underscores its efficacy. Furthermore, a new metric named clamped temporal error (CTE) is proposed to give a more comprehensive evaluation of prediction performance, especially in time-sensitive emergency events of subseconds. Interestingly, as our ESP features can be described in human-readable language naturally, the application of integrating into ChatGPT also shows huge potential. The ESP-dataset and all benchmarks are released at \url{https://dingrui-wang.github.io/ESP-Dataset/}.
\end{abstract}


%
\section{Introduction}
Autonomous driving (AD) is attracting significant attention as well as a large amount of investment and resources. However, the business success of AD is still stuck at level 2/3. The driverless solution, \ie  the level-4 AD, has slowed its pace down towards mass production as safety must be verified with zero-tolerance solidly. To improve the safety of the AD system, one of the most key technologies is prediction. An acute yet reliable algorithm to predict the future states of the surrounding traffic participants is the cornerstone of safe decision and driving control.  

The rapid advance of prediction, especially powered by artificial intelligence techniques, has been witnessed in recent years. The seed work of the deep learning-based prediction algorithm is LSTM\cite{hochreiter1997long, altche2017lstm}. It encodes the states, \eg the velocity and position of traffic participants, in the past few seconds to estimate the future states. We name such historical states based method to \emph{introspective prediction}, which heavily relies on information in its own right. However, such introspective prediction methods suffer from multi-agent interaction. The pioneer work taking the influence of other traffic participants into account is social LSTM\cite{hou2019interactive}. An inspiring work in this line of research is VectorNet\cite{gao2020vectornet}, which encodes the interaction between not only the ego with the agents but also the ego with HDMap, the static environment information. Such lightweight HDMap vectorization technique brings huge performance improvement \wrt the displacement error of predicted trajectories.

\begin{figure}[t]
\centering
\includegraphics[width=1.0\columnwidth]{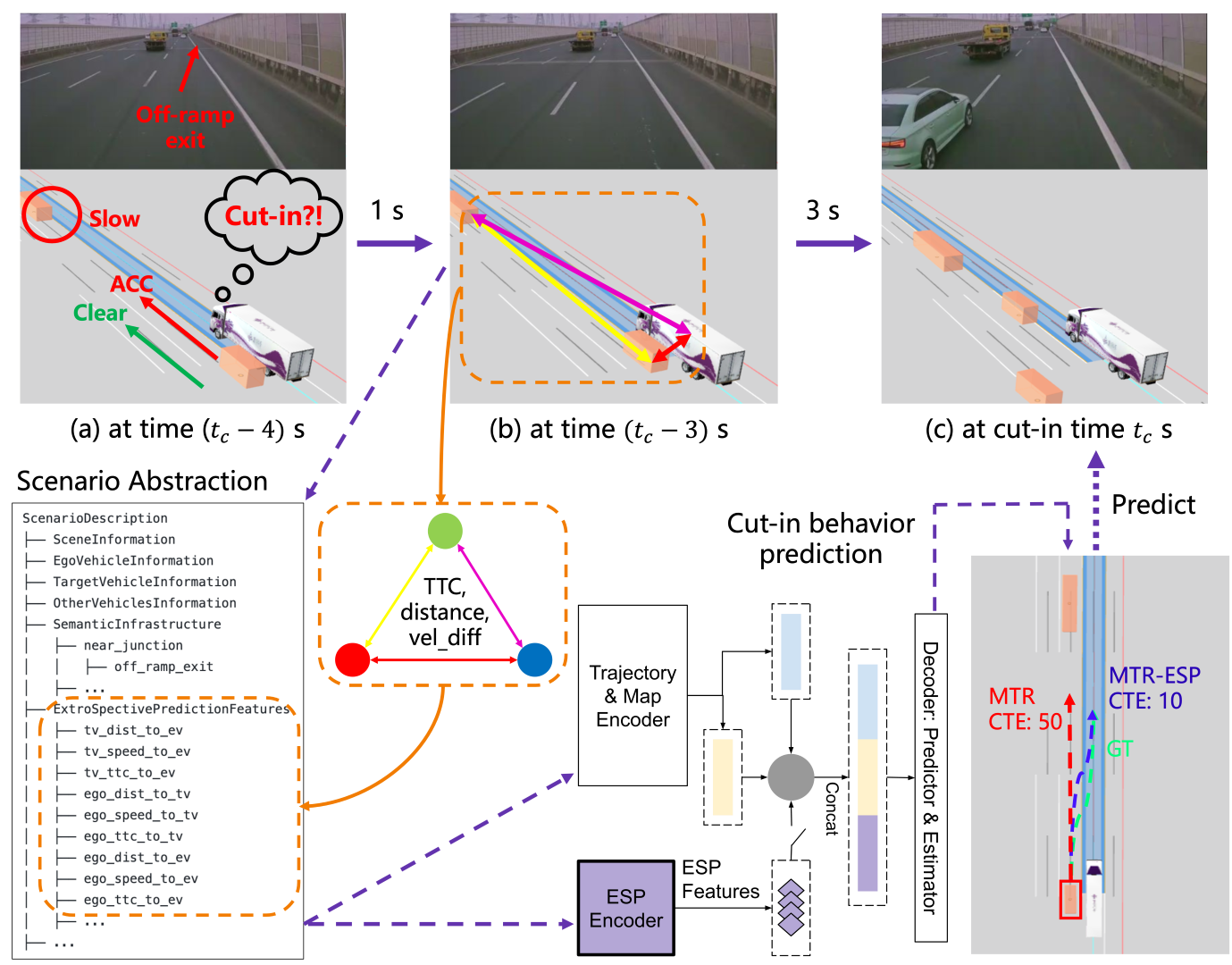}
\vspace{0.05in}
\caption{A real emergency scenario with a sedan dangerously cut-in in front of the AD truck on the highway. At the time in (a), human drivers foretell sedan's behavior by interpreting \emph{extrospective} cues: 1) \emph{[observe]} a high-speed accelerating (\textbf{ACC}) sedan approaching a \textbf{Slow} front-blocking truck, \emph{[predict]} high potential left/right lane change and low possibility of hard brake for the sedan, 2) \emph{[observe]} left lane of the sedan is \textbf{Clear}, \emph{[predict]} left lane change will not happen as it can be done at anytime earlier with lower risk. 3) \emph{[observe]} an \textbf{Off-ramp exit} in about 200 meters, \emph{[predict]} likely to force cut-in into far-right lane to catch exit. Note the sedan did exit the highway as expected in this case. The MTR method predicts the behavior at (c). While the ESP encoder can absorb the \emph{extrospective} cues to predict the cut-in event in advance as shown in (b).}
\label{fig:teaser}
\end{figure}

\begin{table*}[t] 

\caption{A comprehensive comparison of motion prediction datasets.}
\centering
\begin{tabular}%
{c|p{0.4cm}p{0.8cm}p{1cm}p{0.8cm}p{0.6cm}cp{1.2cm} p{0.8cm}p{1cm}p{1cm}p{1.5cm} }
\toprule 

Dataset                             &Year      & Segments      &Time horizon    &Sampling rate  &Boxes  & Distance     & Density of aggressive behavior    & Semantic map    & Highway    & Offline perception   & Semantic environment information     \\
\hline 
\rule{0pt}{0.3cm}Lyft 
\cite{kesten2019lyft}                &2019      &170k           &5s              &10 Hz          &2D     &10 km         &low                                 &{}            &{}             &{}                   &{}               \\ 
\rule{0pt}{0.25cm}TrafficPredict 
\cite{ma2019trafficpredict}          &2019      &{-}           &{-}              &10 Hz         &{-}     &{-}           &high                             &{}     &{}             &{}                   &{}            \\
\rule{0pt}{0.25cm}NuScenes 
\cite{caesar2020nuscenes}           &2020      &1k             &6s              &2 Hz           &3D     &{-}           &low                                 &{}            &{}             &{}                   &{}            \\
\rule{0pt}{0.25cm}Argoverse 
\cite{chang2019argoverse}           &2019      &324k           &3s              &10 Hz          &None   &290 km        &low                                &\checkmark     &{}             &{}                   &{}             \\
\rule{0pt}{0.25cm}INTERACTIONS 
\cite{zhan2019interaction}          &2019      &{-}            &3s              &10 Hz          &2D     &{-}           &high                               &\checkmark     &\checkmark     &\checkmark           &{}    \\
\rule{0pt}{0.25cm}WOMD 
\cite{ettinger2021large}            &2021      &104k           &8s              &10 Hz          &3D     &1750 km       &low                                &\checkmark     &\checkmark     &\checkmark           &{}  \\
\rule{0pt}{0.25cm}ESP-Dataset              &2023      &120k           &5s              &10 Hz          &2D     &2100 km       &very high                          &\checkmark     &\checkmark     &\checkmark           &\checkmark    \\
\bottomrule 

\end{tabular}
\vspace{0.1in}
\label{tab:comparison_between_datasets}
\end{table*}

Even with the social interaction encoding, the SOTA prediction methods~\cite{shi2022motion, zhao2021tnt} are still not as intelligent as a human and always fail when facing complex or emergency scenarios, which requires a deep understanding of the environment and multiple lines of reasoning. Fig.~\ref{fig:teaser} shows a real-world \emph{emergency cut-in} case from our AD fleet. The human driver can predict the emergency cut-in event at the most early timing shown in Fig.~\ref{fig:teaser} (a). The advanced MTR method~\cite{shi2022motion} only yields the prediction when lateral movement is observable at the timing shown in Fig.~\ref{fig:teaser} (c).

The lesson learned from this real case is the huge gap between human extrospective cue understanding and reasoning ability and the performance of current prediction algorithms. Rethinking the MTR/TNT and related works \cite{gao2020vectornet, li2019grip++, sheng2022graph, wang2023ganet}, even attention mechanism incorporating with local/global graph is well-designed to encode historical state and HDMap information, the global attention on extrospective cue and corresponding inference ability is really weak. 
Among the reasons is the deficiency of the dataset focusing on rare and difficult-to-predict scenarios with rich but often-overlooked environmental information. Furthermore, a more proper metric than Final Displacement Error (FDE) and Average Displacement Error (ADE), as well as powerful baselines should be developed for those challenge scenarios.

In this paper, we tackle the problem of long-term prediction in emergency scenarios, where internal states in history are inconspicuous but the external environment provides effective cues for reasoning in the best time window ahead of the emergency event. We name it as the extrospective problem (ESP). Due to the lack of related datasets, we built the ESP-Dataset for challenging scenarios with emergency events. Moreover, the dataset is collected and labeled for diverse scenarios over 2k+ kilometers, with novel and unique semantic environment information for extrospective prediction provided. Furthermore, a new metric named Clamped Temporal Error(CTE) is designed to comprehensively evaluate time-wise prediction performance, which is an important but missing aspect for sub-second-level time-sensitive emergency scenarios. Lastly, ESP feature extraction and network encoder are introduced, which would benefit existing backbones and algorithms seamlessly. 
Back to the emergency case in Fig.~\ref{fig:teaser}, by introducing ESP encoding, MTR with ESP effectively forecasts the cut-in three seconds in advance than MTR shown in Fig.~\ref{fig:teaser} (b). One more interesting thing is that our ESP features can be described in human-readable language naturally, and we have already integrated it into large language models such as GPT~\cite{ouyang2022training} .

The contributions of the paper are summarized below: 

\begin{itemize}
  \item The ESP-Dataset with semantic environment information is collected over 2k+ kilometers focusing on emergency-event-based challenging scenarios. 

  \item A new metric named CTE is proposed for comprehensive evaluation of prediction performance in time-sensitive emergency scenarios. 
  
  \item ESP feature extraction and network encoder are introduced, which can be used to enhance existing backbones/algorithms seamlessly.
\end{itemize}
\section{Related Work}

\subsection{Motion prediction datasets}
In the last decade, significant advancements in Autonomous Vehicle (AV) have been propelled by the availability of diverse datasets for understanding scenes. This journey began with 2D annotated datasets (such as CamVid\cite{brostow2008segmentation} and Apolloscape\cite{huang2019apolloscape}), which evolved into multimodal datasets (like KITTI\cite{geiger2012we} and KAIST\cite{choi2018kaist}), incorporating images, range sensor data (lidars, radars), and GPS/IMU data~\etc. However, these datasets primarily focus on \emph{introspective} cues, as they only consider historical information, neglecting semantic details. 
In addition, numerous other motion prediction datasets also have been established including the Stanford Drone Dataset\cite{robicquet2016learning}, Town Center\cite{benfold2011stable}, NGSIM\cite{coifman2017critical}, ETH\cite{pellegrini2009you}, Automatum\cite{spannaus2021automatum}, UCY\cite{lerner2007crowds}, highD\cite{krajewski2018highd}, and exiD\cite{moers2022exid}. It is worth noting that while these datasets offer valuable insights into motion prediction, they primarily focus on specific fixed locations rather than the broader context of dynamic driving environments.
Moving on to large-scale AV datasets, nuScenes\cite{caesar2020nuscenes}, Lyft L5\cite{kesten2019lyft}, TrafficPredict\cite{ma2019trafficpredict}, Argoverse\cite{chang2019argoverse}, INTERACTION\cite{zhan2019interaction}, and Waymo Open Motion dataset (WOMD)\cite{ettinger2021large} have been made available to the public. The INTERACTION dataset selects particular driving locations (e.g., roundabouts) to emphasize interactive complexity, while WOMD aims to jointly predict motion behavior. As previously mentioned, understanding context requires considering semantic information related to \emph{extrospective} aspect. TrafficPredict, Argoverse, INTERACTION, and WOMD provide HD maps rich in semantics. Nevertheless, these datasets can only provide \emph{introspective} cues to prediction models, neglecting \emph{extrospective} factors.
NuScenes and Lyft L5 dataset differ in that they do not focus on interactive driving scenarios\cite{caesar2020nuscenes}. An overall comparison of various AV datasets is presented in Table~\ref{tab:comparison_between_datasets}.

\subsection{Driving behavior prediction}
Theory of Mind (ToM), the ability to understand the mental states of others, is a primary reason humans can successfully negotiate traffic on a highway onramp~\cite{ivanovic2019trajectron}. As previously mentioned, existing models such as Vectornet~\cite{gao2020vectornet}, TNT~\cite{zhao2021tnt}, Grip++~\cite{li2019grip++}, Graph-based network~\cite{sheng2022graph}, GANet~\cite{wang2023ganet} and MTR \cite{shi2022motion},~\etc tend to focus on \emph{introspective} aspect, which relies heavily on historical data and high-definition maps. Furthermore, their performance is still far from human-level interpretation of semantic environment information, and struggle with complex scenarios that require a deep understanding of the environment through complex reasoning~\cite{karle2022scenario}.
This is precisely why the advancement of current models requires richer and more detailed semantic information. In light of this, we introduce the ESP encoder to demonstrate the seamless integration capabilities of the ESP dataset. Large Language Models (LLMs) such as the chatGPT~\cite{ouyang2022training} and GPT-4~\cite{peng2023instruction} have been increasingly gaining attention recently. And there are recent studies have explored their applications in autonomous driving tasks~\cite{fu2023drive}. Our dataset is also compatible with LLMs regarding input format. 

Concerning the prediction of cut-in behaviors, various attempts have been made such as convolution neural network-based method~\cite{kazemi2018learning} and behavioral probability distribution-based method~\cite{izquierdo2019prevention}. However, existing metrics to evaluate the prediction result such as Time-to-Collision~\cite{moers2022exid}, Final Displacement Error (FDE), Average Displacement Error (ADE), MR (Missing Rate)~\cite{ettinger2021large} and Average Precision~\cite{caesar2020nuscenes} do not adequately assess the quality of cut-in predictions. Specifically, traditional metrics evaluate cut-in predictions based on trajectory rather than the precise cut-in moment. Consequently, current metrics emphasize spatial information while neglecting temporal aspects during evaluation.

\begin{figure}[h]
\centering
\includegraphics[width=0.99\columnwidth]{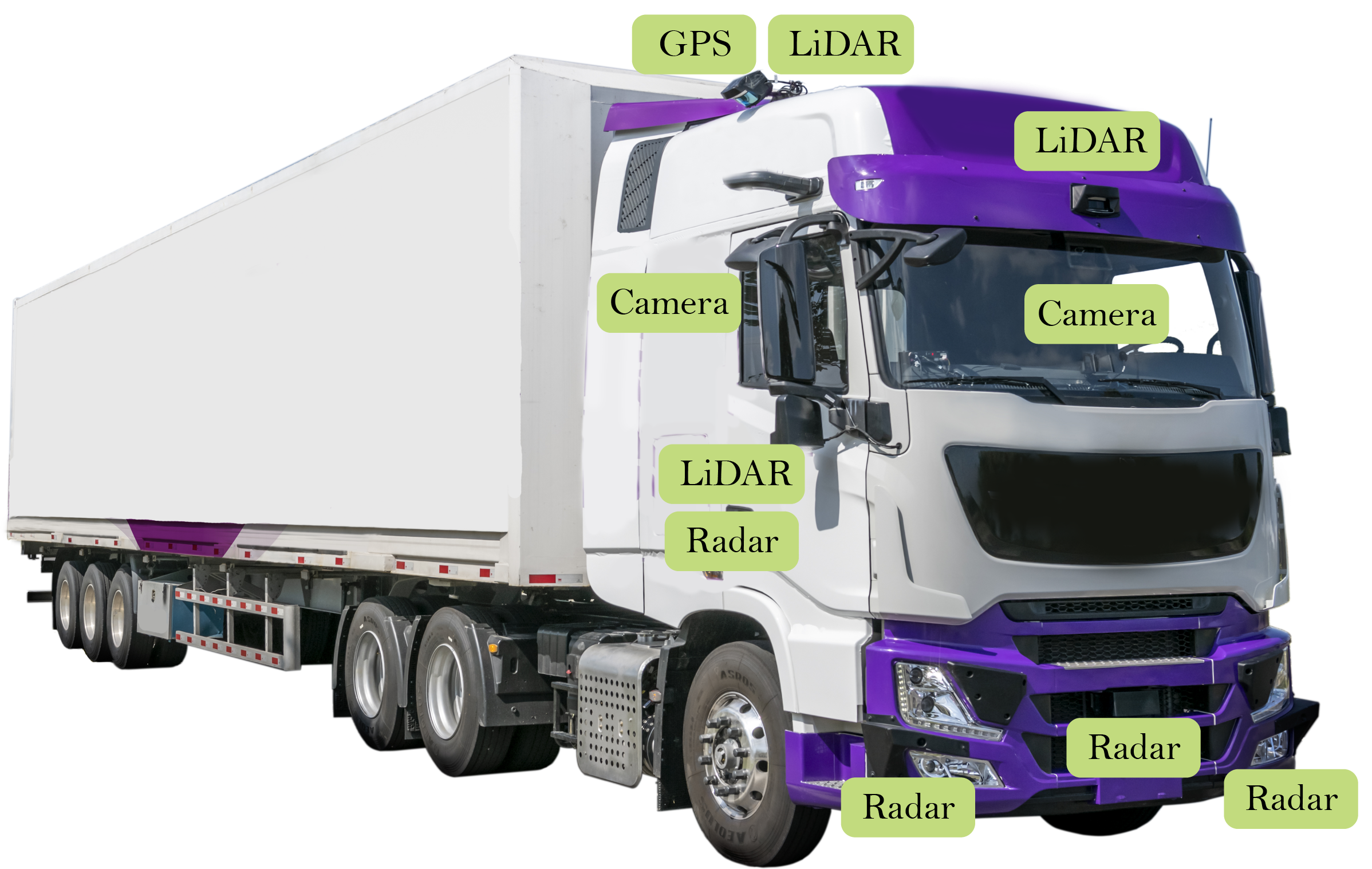}
\caption{Sensor setup for the ESP data collection platform involves the Inceptio autonomous truck, which is equipped with 5 LiDARs, 7 cameras, 7 radars, and GPS.}
\label{fig:sensor}
\end{figure}

\section{The ESP-Dataset}
\subsection{Sensor Setup}

We use Inceptio autonomous trucks \cite{wang2022intelligent} to capture data as shown in Fig. \ref{fig:sensor}. The sensor system of the Inceptio autonomous truck includes $5$ LiDARs, $7$ cameras, $7$ radars, and GPS. All five LiDARs are solid-state. One forward long-range LiDAR is above the front windshield, two rear-facing long-range LiDARs are under the rearview mirrors on both sides, and two short-range blind zone LiDARs are on both sides of the roof. Among them, long-range LiDARs can detect up to 200 meters with 120 degrees FOV and short-range LiDARs can scan up to 140 degrees. Three cameras with short, middle, and long ranges are mounted on the top of the windshield. The other four cameras are installed above the truck doors on both sides, and these cameras are forward middle-range cameras and backward fish-eye cameras. Five long-range radars are placed on the bumper with forward, left-forward, right-forward, left-rear, and right-rear views. The two other two long-range radars are mounted under the rearview mirrors. GPS is installed on the roof. 

\subsection{Scenario Description Paradigm}

A complete representation of the agent's operating environment is critical for testing and evaluating autonomous driving models \cite{li2019aads, queiroz2019geoscenario}. Different models use various representations to describe the driving scenario. In MDP and MCTS \cite{bai2015intention, wei2011point}, it's a state space with vehicle position, velocity, and actions. Lev (Lane-based planning) \cite{wang2022ltp} uses lane information while Free-space velocity (FV) \cite{claussmann2019review} has no strict lane constraints. Social LSTM \cite{hou2019interactive} uses the trajectories of multiple agents. For Graph-based methods \cite{li2019grip++, sheng2022graph}, agents are nodes, and interactions are edges. Moreover, there are models \cite{shi2022motion, zhao2021tnt, wang2023ganet} that focus on using agent trajectories and map polylines as the encoder's input. However, previous models lack a semantics-oriented \emph{extrospective} understanding of scenario descriptions. Large Language Models (LLMs) \cite{wei2022emergent, fu2023drive} are capable of absorbing natural language-oriented information and providing comprehensive insights. Similarly, a driving scenario can be organized in a format that incorporates natural language information.

\begin{figure}[t]
\centering
\includegraphics[width=0.99\columnwidth]{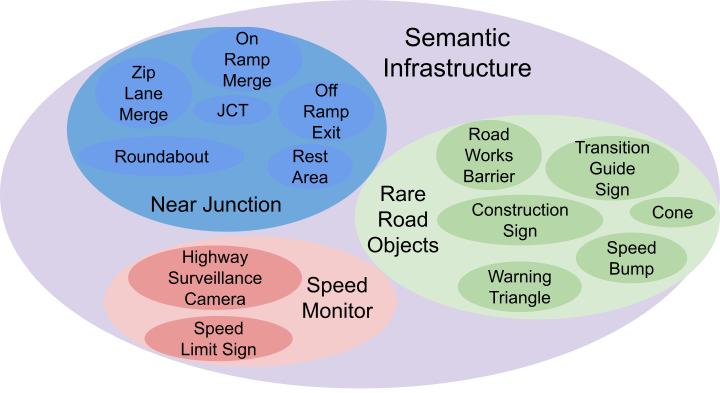}
\caption{Organization of Semantic Infrastructure. The figure illustrates the organization of Semantic Infrastructure, which comprises three extrospective components: speed monitoring systems, junctions, and rare road objects.}
\label{fig:semantic_infra}
\end{figure}

Inspired by this, to describe the scenario in a complete manner, the paradigm of the scenario description to build the dataset includes the following components.

\begin{itemize}[left=0em]
  \item \textbf{Scene} covers diverse attributes like lane type, weather conditions, total vehicles in scope, etc.
  \item \textbf{Ego vehicle}'s lane location together with its historical velocity and trajectory data are given.
  \item \textbf{Target vehicle} is the object for which future behavior needs to be predicted. This section includes information presented in a format similar to that of the ego vehicle, with the distinction that it also includes the ground truth of its future behavior.
  \item \textbf{Relative Interaction Vehicles} describes surrounding vehicles in a format similar to that of the ego vehicle.
  \item \textbf{Semantic Infrastructure} encompasses three \emph{extrospective} components: speed monitoring systems, junctions and rare road objects. The detailed framework of the semantic infrastructure is illustrated in Fig.\ref{fig:semantic_infra}.
  \item \textbf{\emph{Extrospective} Features} contains different features related to the distance and relative longitudinal velocity between different agents in an ESP token.
\end{itemize}

\begin{figure}[h]
\centering
\includegraphics[width=0.99\columnwidth]{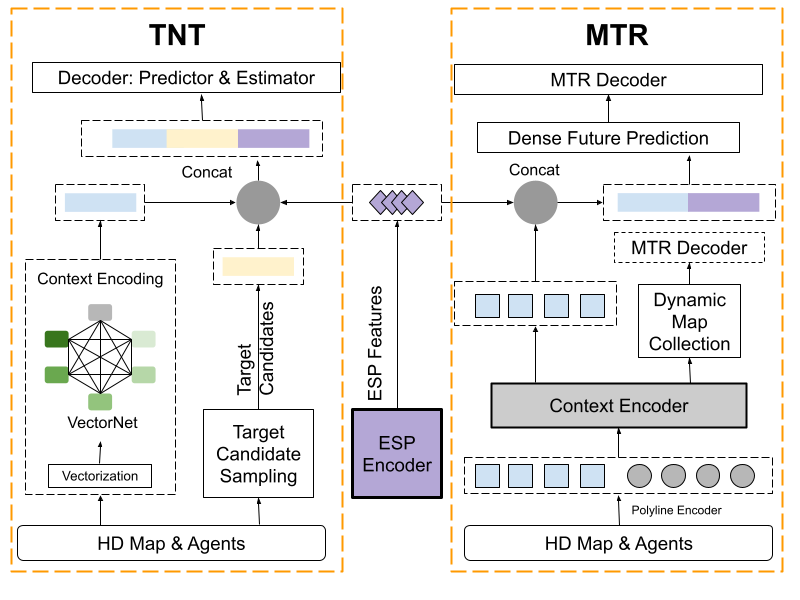}
\caption{Seamless Integration of ESP Features with Motion Prediction Models. The ESP plugin seamlessly integrates with widely used encoder-decoder models such as TNT and MTR. As depicted, ESP enhances existing features through straightforward concatenation, leading to a transformative advancement in motion prediction.}
\label{fig:esp_encoder}
\end{figure}

\subsection{Easy plug-in ESP Encoder}

We propose to extract ESP features with a simple network encoder. Even though the encoder in our case is as simple as a one-layer standardized LSTM module, the improvement to the performance of the SOTA models is already considerable. The related experiment results will be shown in section \ref{sec:applications}. And the results have shown the easy plug-in property of the ESP encoder which can also enhance existing backbones/algorithms seamlessly. As depicted in Fig. \ref{fig:esp_encoder}, the result generated by the ESP encoder is directly concatenated to the original encoder output of the model. The only aspect to be mindful of is to account for the input dimension of the model's decoder.

\subsection{Data mining}

\begin{figure}[t]
\centering
\includegraphics[width=1.0\columnwidth]{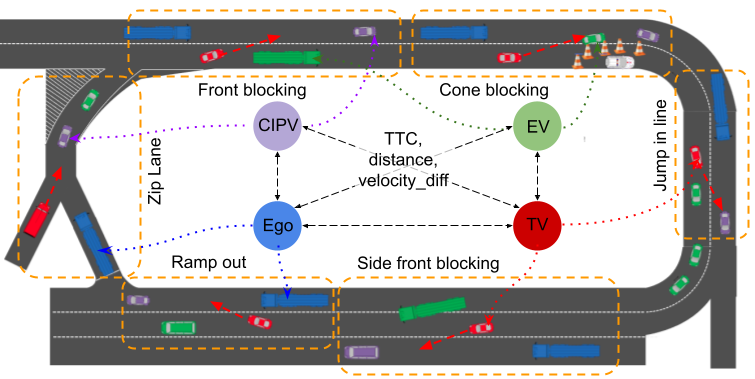}
\caption{ESP Token Types - Representation of Ego (Ego vehicle), CIPV (Vehicle in front of ego), EV (Environmental vehicle), and TV (Target vehicle) within each time frame. Scenario types are determined based on rule-based criteria.}
\label{fig:token_types}
\end{figure}

As depicted in Fig. \ref{fig:token_types}, collected scenarios encompassed various interactions such as merges, lane changes, ramp out, cone block, and zip lane. Interesting scenarios were mined token by token and based on different spatial-temporal criteria which are performed every three frames, using the current frame at time $t_0$, the historical frame at time $t_{-3}$, and the future frame at time $t_3$ as a base for detection. 

An example of the mining detection criteria for front-blocking scenarios is: there is a slow-moving vehicle agent A ahead, and the Time-to-Collision is less than 5 seconds. In the next 5 seconds, if the ego vehicle's minimum deceleration is heavier than a threshold (\eg -0.9 m/s\textsuperscript{2}
) or the average deceleration is heavier than a threshold (\eg -0.5 m/s\textsuperscript{2}
), and the preceding vehicle either performs a cut-in maneuver or is close to the ego vehicle's dangerous zone.

\begin{figure}[h]
\centering
\includegraphics[width=1.0\columnwidth]{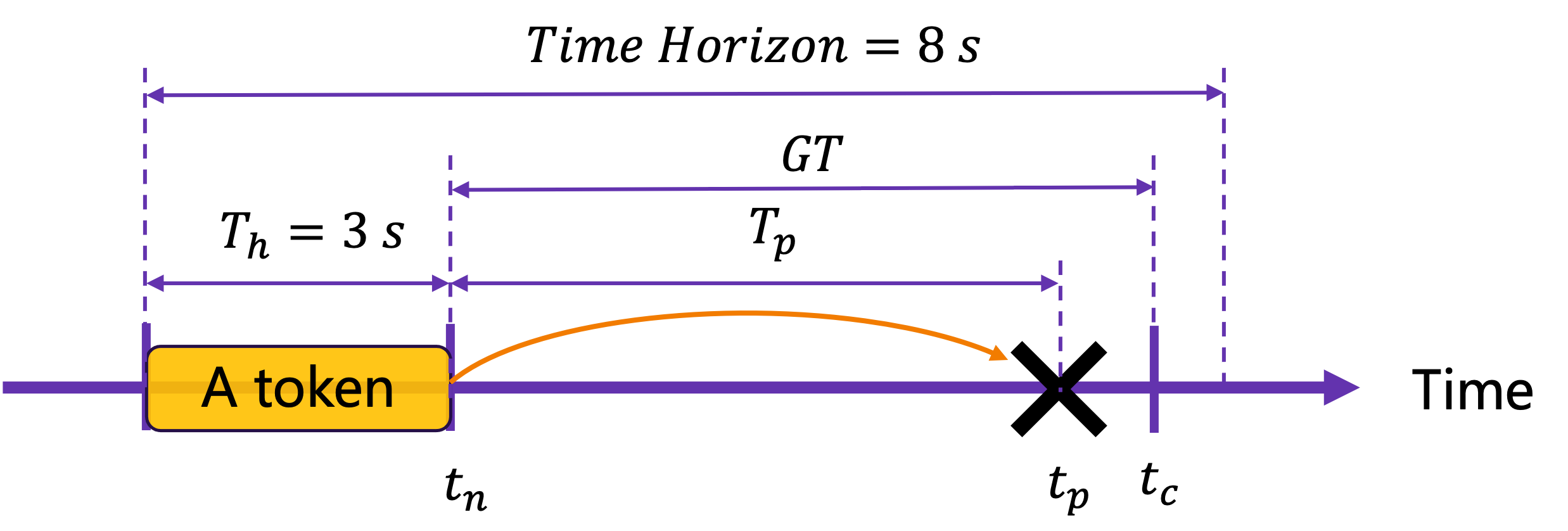}
\caption{This figure depicts the framework of a token. The token spans a history of three seconds and includes a ground truth trajectory for the target vehicle to be predicted over five seconds. Therefore, the overall time horizon is eight seconds. Here, $t_n$ represents the current time stamp, $t_p$ is the predicted time stamp, and $t_c$ denotes the ground truth cut-in moment.}
\label{fig:token_illustration}
\end{figure}

Once these conditions are satisfied, a token for the abstraction of a scenario will be extracted. The mined token result is illustrated in Fig. \ref{fig:token_illustration}, the token spans over a history of three seconds and includes a ground truth trajectory for the
target vehicle to be predicted over five seconds. Therefore, the overall time horizon is eight seconds.

The map data covers the Shanghai-Jinan and Shanghai-Quanzhou highways, as well as the Shanghai Outer Ring Road. We have a representation of lane features including the lane centerlines, lane boundary lines, and road edges.

\subsection{Statistics}
\begin{figure}[h]
\includegraphics[width=1.0\columnwidth]{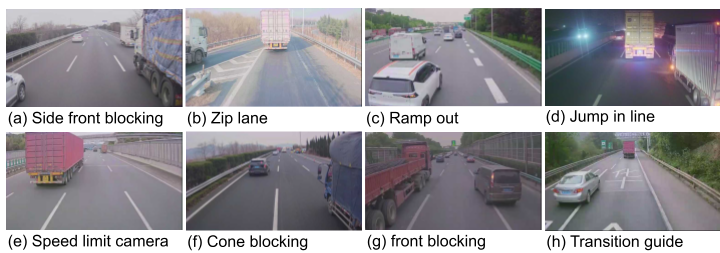}
\caption{Example scenarios from the ESP-dataset captured by the front and side cameras of our ego-truck.}
\label{fig:data_scenarios}
\end{figure}

ESP covers more than 110,000 tokens on highways with a total length of around 2100 km. The annotation frequency is 10Hz. We organize the data by frames, with each frame containing all the traffic agents' IDs, categories, positions, and bounding boxes. We divide the dataset into training, validation, and testing sets by 8:1:1. Fig. ~\ref{fig:data_scenarios} provides 8 typical scenarios sampled in the ESP-dataset. It can be seen that the traffic in our dataset features on challenging scenario that has high density, the surrounding areas of the ego-truck are occupied by heterogeneous vehicles, such as cars and trucks. 

\section{Task and Metric}
\subsection{Cut-in Evaluation Dilemma}

\begin{figure}[ht]
\includegraphics[width=1.0\columnwidth]{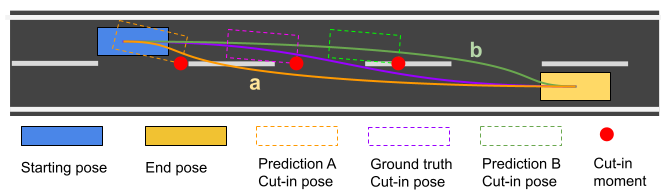}
\caption{Illustration of the necessity to consider CTE metric in the assessment of cut-in behavior prediction.}
\label{fig:CTE_illustration}
\end{figure}

Regarding driving behavior, with a specific focus on cut-in behavior, existing models primarily centered on trajectory prediction can also provide predictions for cut-in behavior. This can be accomplished by superimposing the bounding box onto the predicted trajectory while considering the heading for each point along the trajectory, the initial intersection point between the bounding box and the lane's polyline can be identified. Despite the straightforward conversion of trajectory prediction into behavior prediction, assessing the prediction outcomes based on current metrics such as Final Displacement Error (FDE), Average Displacement Error (ADE) and MR (Missing Rate) poses a non-trivial challenge. The reason behind the challenge is that these metrics focus on overall trajectory performance without considering temporal details. For example, in the Fig.~\ref{fig:CTE_illustration}, trajectories labeled ``a'' and ``b'' would yield the same results with these metrics regarding the ground truth in between, despite ``a'' showing an earlier lane change. These observations highlight the limitations of current metrics and emphasize the need to capture the temporal aspects of prediction behaviors for a more comprehensive evaluation.

\subsection{Metric}


We introduce the Clamped Temporal Error (\textbf{\textit{CTE}}) to measure the difference between predicted and actual behavior times. For each scenario token $S$ that we evaluate, our model generates $K$ potential predictions, denoted as $P_k$, $k \in 1 \ldots K$. Each prediction $P_k$ is related to a trajectory $s_k = \{s_{a,t}\}_{t=1:T, a=1:A}$ for $T$ future time steps for $A$ agents. Similarly, the ground truth is denoted as $\hat{s} = \{\hat{s}_{a,t}\}$. The individual object prediction task becomes a special case of this formulation where each joint prediction contains only a single agent $A = 1$.
The \textbf{\textit{minCTE}} is computed as formulated in the equation below,
\begin{equation}
\textbf{\textit{minCTE}} = \min_k \sum_{a=1}^{A} \min(||LaMT(\hat{s}_{a,t}) - LaMT(s_{a,t}^{k})|| , t_u)
\end{equation} 
where $LaMT$ represents the function for calculating Lane Match Time by determining the intersection point between the input trajectory and the polylines of nearby lanes \wrt the vehicle's heading and boundary box. The term $t_u$ refers to the upper threshold to clamp the time difference.

\begin{table}[h]
    \caption{Ablation study on ESP features using TNT, re-trained TNT, and MTR with ESP encoder.}
    \centering
    
    \begin{tabular}{p{1.6cm}p{0.7cm}p{0.7cm}p{0.7cm}p{0.7cm}p{0.6cm}p{0.6cm}}
        \toprule
        \rule{0pt}{0.2cm} Model                             & minFDE        & minADE        & minCTE        & Precis.     & Recall        & Acc. \\
        \hline
        \rule{0pt}{0.3cm} \textbf{TNT-ESP}& \textbf{\emph{4.33}}   & \textbf{\emph{2.00}}   & \textbf{\emph{0.71}}   & \textbf{\emph{0.75}}   & \textbf{\emph{0.83}}   & \textbf{\emph{0.76}} \\
        \rule{0pt}{0.2cm} w/o tv-ev               & \emph{4.89}   & \emph{2.21}   & \emph{0.74}   & \emph{0.73}   & \emph{0.83}   & \emph{0.75} \\
        \rule{0pt}{0.2cm} w/o tv-cipv             & \emph{5.22}   & \emph{2.32}   & \emph{0.75}   & \emph{0.75}   & \emph{0.80}   & \emph{0.75} \\
        \rule{0pt}{0.2cm} w/o ego-tv              & \emph{5.88}   & \emph{2.59}   & \emph{0.82}   & \emph{0.72}   & \emph{0.78}   & \emph{0.72} \\
        \rule{0pt}{0.2cm} w/o ego-cipv            & \emph{4.87}   & \emph{2.20}   & \emph{0.74}   & \emph{0.73}   & \emph{0.83}   & \emph{0.74} \\
        \rule{0pt}{0.2cm} w/o ego-ev              & \emph{5.23}   & \emph{2.33}   & \emph{0.75}   & \emph{0.72}   & \emph{0.82}   & \emph{0.74} \\
        \rule{0pt}{0.2cm} TNT-base  & \emph{5.21}   & \emph{2.33}   & \emph{0.75}   & \emph{0.74}   & \emph{0.81}   & \emph{0.74} \\
        \hline
        \multicolumn{2}{l}{\rule{0pt}{0.3cm} TNT-ESP (retrained)}&  &    &    &    &  \\
        \rule{0pt}{0.2cm} w/o tv-ev              & \emph{4.46}   & \emph{2.06}   & \emph{0.75}   & \emph{0.72}   & \emph{0.84}   & \emph{0.74} \\
        \rule{0pt}{0.2cm} w/o tv-cipv             & \emph{4.45}   & \emph{2.06}   & \emph{0.73}   & \emph{0.73}   & \emph{0.85}   & \emph{0.75} \\
        \rule{0pt}{0.2cm} w/o ego-tv              & \emph{4.46}   & \emph{2.05}   & \emph{0.73}   & \emph{0.73}   & \emph{0.84}   & \emph{0.75} \\
        \rule{0pt}{0.2cm} w/o ego-cipv           & \emph{4.52}   & \emph{2.08}   & \emph{0.73}   & \emph{0.73}   & \emph{0.84}   & \emph{0.75} \\
        \rule{0pt}{0.2cm} w/o ego-ev            & \emph{4.44}   & \emph{2.04}   & \emph{0.75}   & \emph{0.72}   & \emph{0.85}   & \emph{0.74} \\
        \rule{0pt}{0.2cm} TNT-base  & \emph{4.52}   & \emph{2.07}   & \emph{0.74}   & \emph{0.74}   & \emph{0.83}   & \emph{0.75} \\

        \hline
        \rule{0pt}{0.3cm} \textbf{MTR-ESP}             & \textbf{\emph{1.04}}   & \textbf{\emph{0.56}}   & \textbf{\emph{0.23}}   & \textbf{\emph{0.72}}   & \textbf{\emph{0.955}}   & \textbf{\emph{0.932}} \\
        \rule{0pt}{0.2cm} MTR-base                & \emph{1.06}   & \emph{0.58}   & \emph{0.24}   & \emph{0.70}   & \emph{0.953}   & \emph{0.930} \\
        \bottomrule 
    \end{tabular}
    \vspace{-0.10in}
    \label{tab:ablation_study_tnt}
\end{table}

\section{ESP Applications}\label{sec:applications}

\subsection{The Effectiveness of ESP Encoder }
\begin{figure}[t]
\centering
\includegraphics[width=1.0\columnwidth]{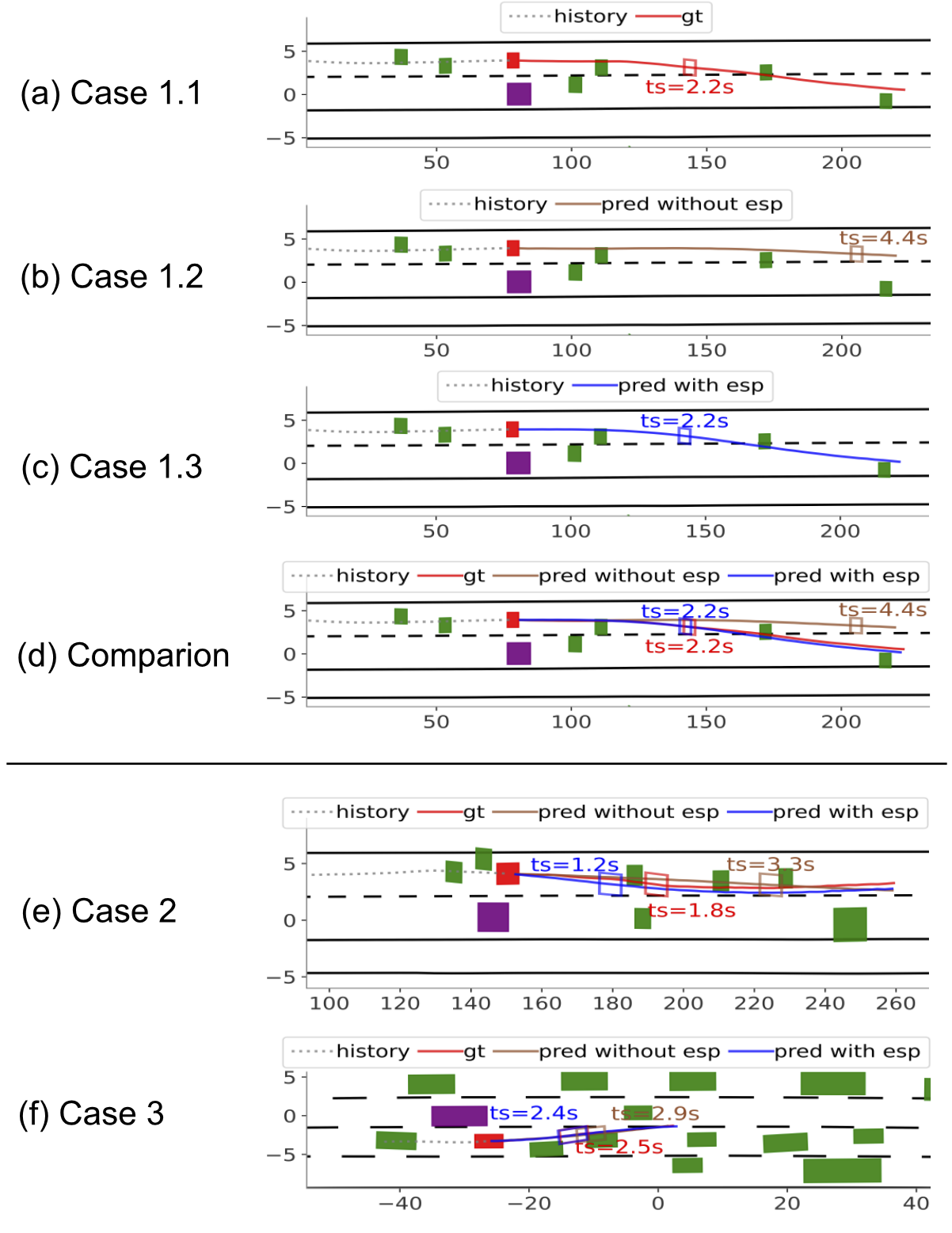}
\caption{MTR-base and MTR-ESP prediction performance comparison. Case 1.1 to Case 1.4 are the ground truth trajectory for the cut-in behavior, the prediction results of the MTR-base model without ESP features, the prediction results of the MTR model with the ESP encoder, and a combined plot respectively.
}
\label{fig:prediction_result}
\end{figure}

The ESP encoding has demonstrated its significant value by producing rapid and discernible improvements in model performance utilizing the ESP features from the ESP-dataset. As shown in Table \ref{tab:ablation_study_tnt}, our initial experiment with TNT \cite{zhao2021tnt} serves to elucidate the distinct influences exerted by each component of the ESP encoder. It is evident that the ESP encoder substantially enhances performance in nearly all aspects.
Furthermore, our ablation study on TNT, retrained with only different partial ESP features, reveals that different segments of the ESP features contribute to varying degrees of influence on the outcome. Additionally, in our experiment involving MTR \cite{shi2022motion}, the results demonstrate that the ESP encoder can still enhance the state-of-the-art (SOTA) baseline's performance considerably. In Fig.~\ref{fig:prediction_result}, three cases are presented to demonstrate the substantial improvement in the model's performance for cut-in behavior prediction achieved through MTR with the use of the ESP encoder. As in Case 1.1, the ground truth cut-in moment occurs at 2.2 seconds, whereas the MTR-base model predicts it at 4.4 seconds (Case 1.2). On the contrary, the MTR-ESP model provides a more accurate prediction of 2.2 seconds (Case 1.3).



\begin{figure}[ht]
    \centering
    \includegraphics[width=1.0\linewidth]{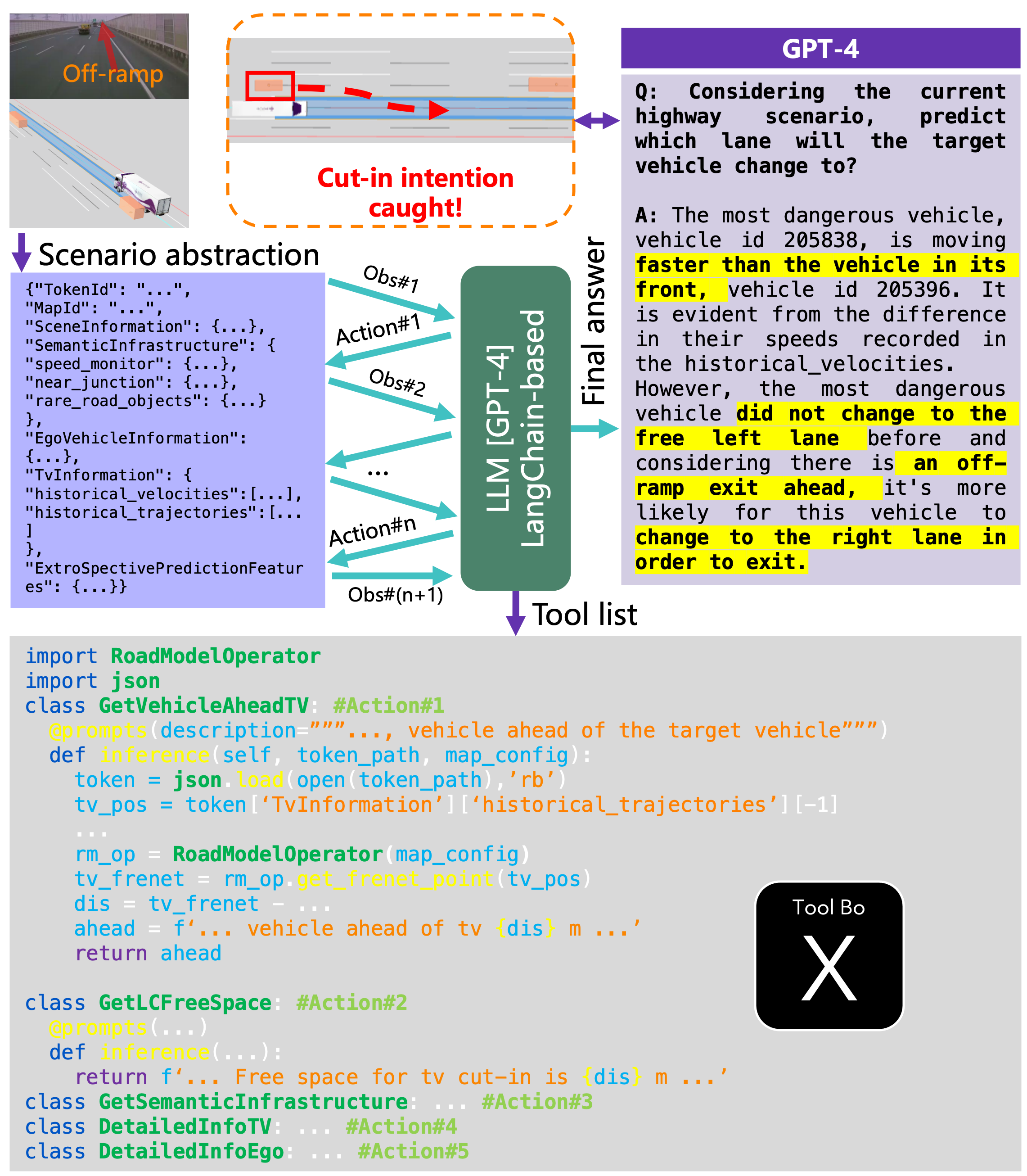}
    \caption{Demonstration of highway scenario processing using the LangChain-based pipeline with GPT-4. GPT-4 initiates with the 'GetVehicleAheadTV' tool and receives an observation indicating the presence of a small vehicle ahead of the target vehicle (TV). Following further inferences, the model ultimately predicts the cut-in intention of the target vehicle in advance. Further, the explanations provided by the model are similar to human thought processes.}
    \label{fig:gpt_demo}
\end{figure}

\subsection{Friendly to LLM}
The GPT model \cite{peng2023instruction} has gained widespread attention in recent months. The emergence of ChatGPT \cite{ouyang2022training} has been captivating the world. Naturally, this leads to the question of how GPT can contribute to the field of motion prediction. In this paper, we explore one potential application of the ESP-dataset: utilizing its tokens directly as input for a Large Language Model. We specifically evaluate its performance within the context of the widely endorsed LangChain framework \cite{chase2022langchain}, as recent studies have advocated for its application in prediction tasks \cite{fu2023drive}. The model pipeline commences by converting the highway scenario into a standardized format and feeding it into GPT-4 through the LangChain interface. Subsequently, the LLM provides insights through multiple action-observation pairs. Actions are defined within the ``toolbox". Once GPT believes that it has gathered enough observational information, it provides the final answer. The driving scenario mentioned in the introduction section goes through this LangChain-based pipeline, with details presented in Fig.\ref{fig:gpt_demo}.

\section{Conclusion and Future Work}
In this paper, we addressed the critical challenge of long-term prediction in autonomous driving, with a focus on emergency cut-in scenarios where semantic environmental cues are pivotal. We introduced the ESP problem, curated the ESP-Dataset enriched with semantic environment information, and introduced the Clamped Temporal Error (CTE) metric for time-sensitive emergency scenario assessment. Our ESP feature extraction with the ESP encoder significantly boosted existing prediction methods, particularly in complex interaction scenarios, as evidenced by TNT and MTR model experiments. Furthermore, we unveiled the potential of incorporating ESP features into large language models like GPT to make better predictions by using \emph{extrospective} cues.

In the future, we plan to make the ESP-dataset available to the research community and also continue expanding the ESP-Dataset by including more diverse and challenging scenarios. Additionally, our research will focus on developing advanced ESP encoders that incorporate causal reasoning techniques for further improvement. Furthermore, fine-tuning LLMs with ESP features, aiming to improve scenario understanding for autonomous vehicles, is an interesting direction.


{\small
\bibliographystyle{icra}
\bibliography{egbib}
}
\end{document}